\documentclass[manuscript,nonacm]{acmart}
\AtBeginDocument{%
  }

\setcopyright{acmlicensed}
\copyrightyear{2026}
\acmYear{2026}
\acmDOI{tbd}
\acmConference[ICPM '27]{International Conference on Process Mining}{February 08--12,  2027}{Rende, Italy}
\setcopyright{none}

\usepackage{booktabs}
\usepackage{multirow}
\usepackage{graphicx}
\usepackage{wrapfig}
\usepackage{array}
\usepackage{paralist}
\usepackage{amsmath}
\usepackage{bm}
\usepackage{enumitem}
\usepackage{subcaption}

\newcommand{\mypar}[1]{\vspace{0.1em}\noindent\textbf{#1.}}
\newcommand{\myparit}[1]{\vspace{0.1em}\noindent\textit{#1.}}

\usepackage[table]{xcolor}

\newcommand{\pos}[1]{\cellcolor{blue!12}#1}
\newcommand{\nega}[1]{\cellcolor{red!12}#1}

\usepackage{etoolbox}

\begin{document}

\title[Predicting Process-Level Behavior]{Predicting Process-Level Behavior: A Benchmarking Study}

\author{Michael Grohs}
\email{michael.grohs@uni-mannheim.de}
\orcid{0000-0003-2658-8992}
\affiliation{%
  \institution{University of Mannheim}
  \city{Mannheim}
  \state{BW}
  \country{Germany}
}
\author{Lukas Kirchdorfer}
\email{lukas.kirchdorfer@sap.com}
\orcid{0000-0003-4713-9328}
% \affiliation{%
%   \institution{University of Mannheim}
%   \city{Mannheim}
%   \state{BW}
%   \country{Germany}
% }
\affiliation{%
  \institution{SAP Signavio}
  \city{Berlin}
  \country{Germany}
}

\author{Jana-Rebecca Rehse}
\email{rehse@ifi.uzh.ch}
\orcid{0000-0001-5707-6944}
\affiliation{%
  \institution{University of Zurich}
  \city{Zurich}
  \country{Switzerland}
}

\renewcommand{\shortauthors}{Grohs et al.}

\begin{abstract}
% Predictive process monitoring (PPM) typically forecasts the future of individual process instances.  However, process managers typically operate on the level of the whole process, e.g., when assigning staff or calculating KPIs. The question hence is how well those process-level managerial decisions are supported by existing approaches.

%   Predictive process monitoring (PPM) typically forecasts the future of individual process instances, yet many managerial decisions require predictions of aggregate, process-level behavior. In this paper, we investigate whether trace-level PPM approaches can accurately provide such process-level forecasts. We formalize two prediction tasks—concurrent trace prediction and average throughput time prediction—and processatically compare aggregated trace-level predictions with statistical and machine-learning-based time series models, foundation models, and process simulation across synthetic and real-life event logs. Our results show that direct process-level forecasting generally outperforms trace-level PPM and often also requires substantially less training effort. In particular, trace-level models systematically underestimate long-running traces, causing errors to accumulate when aggregated to the process level. These findings demonstrate that models optimized for individual traces are not necessarily suitable for forecasting aggregate process behavior and motivate treating process-level prediction as a distinct predictive process monitoring task.

Predictive process monitoring (PPM) typically forecasts the future of individual process instances, hence operating on trace level.  
Process managers, however, typically operate on process level, e.g., when assigning staff or calculating KPIs. 
The question is how well such process-level managerial decisions are supported by existing approaches, given that PPM research has so far not focused on process-level predictions. 
We answer this question through a comprehensive benchmarking study of two process-level prediction tasks: concurrent trace prediction and average throughput time prediction. 
Across synthetic and real-life event logs, we compare statistical and machine-learning-based methods, foundation models, business process simulation, and, importantly, established trace-level PPM approaches, aggregating their predictions to the process-level. 
We assess which modeling paradigms are most suitable for predicting process-level behavior and whether existing trace-level PPM can be effectively repurposed for this task.  
Our results show substantial variation in performance across approaches. 
Notably, aggregated trace-level predictions cause errors to compound on process-level, reducing accuracy and highlighting process-level prediction as a distinct predictive task that warrants dedicated research.
%with trace-level PPM generally underperforming, highlighting process-level prediction as a distinct predictive task that warrants dedicated research.

\end{abstract}

\begin{CCSXML}
<ccs2012>
<concept>
<concept_id>10010405.10010406.10010412.10010415</concept_id>
<concept_desc>Applied computing~Business process monitoring</concept_desc>
<concept_significance>500</concept_significance>
</concept>
</ccs2012>
\end{CCSXML}

\ccsdesc[500]{Applied computing~Business process monitoring}

\keywords{Process Mining, Predictive Process Monitoring, Process-Level Behavior}


\maketitle

% Kontext: process-level predictions, wo die meisten Management Entscheidungen getroffen werden -> Können wir diese Entscheidungen auf Prozess-Ebene unterstützen?

% Die meisten PPM Ansätze sind trace-level; das ist aber ein anderes Problem Feld; wie gut unterstützen trace-level PPM Approaches das im Vergleich zu time-series & ML 

% Um das zu bestimmen machen wir Benchmarking von einer Hand voll Approaches die es gibt plus weitere, die grundsätzlich geeignet wären; 

\section{Introduction}
Predictive process monitoring (PPM) is a branch of process mining that aims to forecast the future behavior of business processes, enabling proactive process management~\cite{ceravolo2024predictive}. %,diFrancescomarino2026predictive}. 
Most PPM approaches thereby operate on trace level, predicting the future of individual process instances~\cite{difrancescomarino2022predictive}. 
This supports trace-level decisions, such as escalation. 
Other process management decisions, however, concern the process as a whole. 
%For example, managers may anticipate future workload to adapt staffing levels or forecast the throughput time to identify emerging performance problems. 
For example, managers may adapt staffing levels to handle increasing workloads or remove activities to reduce throughput time. 
Such decisions require \textit{process-level predictions}: forecasts of aggregate process behavior, captured through key performance indicators (KPIs) such as workload or throughput time~\cite{folino2015prediction,cuzzocrea2018predictive}. 
Whereas trace-level predictions support interventions for individual traces, such process-level predictions can thus support higher-level decisions concerning capacity, resources, and process design.

Despite their practical relevance, process-level predictions are underexplored in PPM research, leaving the suitability of approaches unclear.
%unclear which approaches are most suited for them. 
Some works have shown that aggregate process behavior can be forecast in the form of throughput time~\cite{leribaux2025actor}, constraint violations~\cite{folino2015prediction,cuzzocrea2018predictive}, or work in progress~\cite{mehrdad2024work}, but they each focus on one specific prediction target and a particular model family. 
%thus not assessing several models across prediction tasks and event logs.
%leaving it unclear which approaches are generally suitable for process-level prediction.
Importantly, it is also unclear to what extent existing trace-level PPM approaches can be leveraged for process-level predictions. 
In principle, those approaches can predict process-level KPIs based on collections of traces.
%In practice, however, trace-level approaches have been successfully studied in tasks like remaining time or future event prediction~\cite{ceravolo2024predictive}, but their effectiveness in predicting process-level behavior has not been investigated.
However, the efficacy and accuracy of such aggregated predictions has not yet been investigated.

Against this background, we present a comprehensive benchmark of approaches for predicting process-level behavior.
We consider two complementary prediction tasks: \textit{Concurrent Trace Prediction (CTP)}, forecasting the number of simultaneously active traces, and \textit{Throughput Time Prediction (TTP)}, forecasting the average completion time of traces.
Across synthetic and real-life event logs, we compare a broad spectrum of prediction paradigms for process-level prediction:
\begin{inparaenum}[(1)]
    \item approaches also used in existing work on process-level KPI forecasting from statistical and machine learning (ML) methods, 
    \item recently emerged tabular and time series foundation models, which appear promising for process-level prediction,
    \item process simulation approaches, which provide an alternative means of projecting future behavior, and 
    \item established trace-level PPM approaches, with individual predictions aggregated to the process level. 
\end{inparaenum}
This broad comparison allows us to assess which methods are most suitable for process-level prediction and, in particular, whether existing trace-level approaches can be effectively repurposed for this task.
Our results show that trace-level approaches consistently underperform, mainly because they underestimate long-running traces, causing errors to accumulate at the process level. 
This indicates that trace-level approaches are not necessarily suitable for forecasting process-level behavior. 
Though usually better, the approaches that directly forecast process-level KPIs are also often inaccurate, underscoring the need to treat process-level prediction as a distinct predictive task that requires tailored solutions.

The remainder of the paper is structured as follows:
Section~\ref{sec:rel_work} discusses related work.
Section~\ref{sec:preliminaries} introduces preliminaries and defines two process-level prediction tasks. Then, Section~\ref{sec:eval} describes our experimental setup, followed by a presentation and discussion of the results in Section~\ref{sec:results}. 
Finally, Section~\ref{sec:conclusion} concludes the paper.

\section{Related Work}
\label{sec:rel_work}
Our study relates to several predictive approaches. We first review typical trace-level PPM, before discussing alternatives to forecasting future behavior: process-level performance forecasting, process model forecasting, and process simulation.

\mypar{Trace-level PPM}
Typical PPM approaches predict the future of running traces~\cite{difrancescomarino2022predictive}, in terms of, e.g., upcoming activities or remaining time. 
%In recent years, a large stream of works improved the accuracy of such trace-level predictions. 
We refer to a recent overview existing approaches \cite{ceravolo2024predictive} and only highlight those that we use later, covering relevant prediction tasks (suffix and remaining time prediction) as well as distinct architectures. 
GLSTM uses a generative LSTM network to autoregressively predict the trace suffix~\cite{camargo_DGEN}.
ProcessTransformer (PT) leverages an encoder-only transformer architecture with multi-head self-attention, usable for both suffix and remaining time prediction~\cite{bukhsh2021processtransformer}. 
PGT represents a state-of-the-art remaining time prediction approach based on graph-transformer networks~\cite{PGTNet}.
They differ in architecture and complexity, but are all trained to minimize prediction errors for individual traces. Their predictions can be aggregated to the process level, but this has not yet been studied. %It therefore remains unclear whether trace-level PPM can provide accurate process-level forecasts.

\mypar{Process-level KPI forecasting}
Several existing works directly forecast aggregate process-level KPIs. For instance, throughput time has been predicted based on statistical and ML models, both classical and deep learning, with actor-behavior features improving forecasting performance~\cite{leribaux2025actor}. Others predict the proportion of traces violating aggregate constraints with regression models~\cite{folino2015prediction,cuzzocrea2018predictive}, process-level work-in-progress with convolutional neural networks~\cite{mehrdad2024work}, or inter-case performance indicators like workload on process segments with different neural networks~\cite{denisov2019performance}. Workload has also been forecast from time series of arriving traces using ML approaches~\cite{albertetti2020workload}. 
These works demonstrate that direct process-level prediction is feasible, typically with focus on a single KPI. Most works use ML-based time series forecasting and do not compare themselves to many alternatives. To obtain an overview of which approaches work best, our benchmark includes both statistical and ML methods that were previously applied in process-level forecasting. 
%We also consider process simulation and aggregated trace-level predictions as alternatives, thus studying a broad spectrum of process-level forecasting approaches.

\mypar{Process model forecasting}
Beyond specific KPIs, some approaches predict the future of a process in form of directly-follows relation frequencies using statistical~\cite{de2023process}, deep temporal~\cite{zhou2025process}, or time series foundation models~\cite{yu2026time}. Further work benchmarks univariate and multivariate variants of model forecasting~\cite{yu2025benchmarking} and extends the idea toward logs~\cite{zhou2024event}. 
Process model forecasting shares the process-level perspective of our work where predictions are derived from aggregated signals rather than individual traces, %It also demonstrates the applicability of statistical, ML-based, and foundation-model forecasting approaches to process data. 
but its prediction target differs fundamentally. 
%the evolution of the processes' control-flow is predicted, whereas we focus on process-level KPIs like workload and throughput time. 
Nevertheless, our benchmark also includes the architectures considered here, such as foundation models.

\mypar{Process simulation}
Simulation can support process-level forecasting by learning a simulation model from an event log to generate future process executions. Although simulation is primarily used for what-if analysis, i.e., comparing the effects of alternative process configurations, it can also be applied in an as-is setting to estimate how the process will evolve. Most approaches simulate the arrival and evolution of traces, resulting in simulated logs from which process-level KPIs can be derived. 
Two common simulation paradigms are discrete-event and agent-based simulation, of which SIMOD~\cite{Simod} and AgentSimulator~\cite{KirchdorferBKAS24} are prominent instantiations.
% For example, SIMOD~\cite{Simod} discovers a BPMN model enriched with simulation parameters and executes it using discrete-event simulation. AgentSimulator~\cite{KirchdorferBKAS24}, in contrast, discovers an agent-based simulation model in which process behavior emerges from agent actions and interactions. 
A different approach, called PMSD, discovers a system dynamics model that operates directly on aggregate variables such as work in progress, arrival rates, and completion rates~\cite{pourbafrani2022discovering}. Thus, simulation can provide process-level forecasts either bottom-up from simulated trace-level behavior or directly through process-level models.

% Existing work thus offers accurate trace-level predictions, KPI-specific forecasts enriched with covariates, or forecasts of process structure, but no systematic comparison of which paradigm suits long-horizon, process-level KPI forecasting. In this light, we benchmark the full range of applicable approaches -- traditional PPM, statistical and ML time series models, foundation models, and simulation -- to determine which works best to predict process-level behavior.
Overall, existing work provides trace-level predictions, forecasts of individual process-level KPIs or control-flow behavior, and simulation-based projections. In our benchmark, we compare these four prediction paradigms for two process-level prediction tasks. This enables us to determine which approaches work best and to assess whether the substantial progress in trace-level PPM and process simulation can be repurposed for process-level predictions.
%However, it remains unclear which approaches can best predict process-level KPIs. We address this gap by systematically comparing trace-level PPM with statistical and ML-based time series forecasting, foundation models, and process simulation across two process-level prediction tasks and multiple event logs.

\section{Preliminaries and Prediction Tasks}
\label{sec:preliminaries}

We now introduce preliminaries on event data and process-level KPI time series, and define our two prediction tasks.

\mypar{Event data}
We assume that process executions are recorded in an event log $\mathcal{L}$, defined as a finite set of traces. Each trace  $\sigma = \langle e_{1}, e_{2}, ..., e_{n}\rangle$ represents the temporally ordered sequence of events for a single process instance. An event is a tuple  $e = (c, a, t, r)$, where $c$ denotes the trace ID and $a$ denotes the activity that was executed at time $t$ by some resource $r$. Let $\pi_T$ be a function mapping the event to its timestamp, such that $\pi_T(e) = t$.
For a trace $\sigma=\langle e_1,\ldots,e_n\rangle$, we define its start time as $t_s(\sigma)=\pi_T(e_1)$, its completion time as $t_c(\sigma)=\pi_T(e_n)$, and its throughput time as
$d(\sigma)=t_c(\sigma)-t_s(\sigma)$.

\mypar{Process-level KPI time series}
To represent process-level KPIs over time, we partition the observation period into an ordered sequence of non-overlapping time windows
$\mathcal{W}=\langle W_1,\ldots,W_M\rangle$.
%Here, each window is one calendar day, though other granularities are possible. 
Let
$\kappa$ denote a process-level KPI and
$g_\kappa(\mathcal{L},W_i)$ an aggregation function that computes the value of $\kappa$ for $W_i$ based on traces in $\mathcal{L}$. The resulting time series is defined as
\begin{equation}
K^\kappa
=
\langle k^\kappa_1,\ldots,k^\kappa_M\rangle,
\qquad
k^\kappa_i = g_\kappa(\mathcal{L},W_i).
\label{eq:time_series}
\end{equation}

\mypar{Prediction tasks}
We define two process-level prediction tasks that capture different dimensions of process performance. Since process-level behavior is typically managed via KPIs that are monitored continuously to track how a process evolves over time, we define these prediction tasks as time series forecasting problems: Concurrent Trace Prediction (CTP) forecasts process workload; Throughput Time Prediction (TTP) forecasts time efficiency. 
We first formalize the general setting and then instantiate it for the two tasks.

\myparit{General prediction setting}
For a process-level KPI $\kappa$, we assume that its values have been observed for the first $M$ time windows, yielding the historical KPI time series
$K^\kappa=\langle k^\kappa_1,\ldots,k^\kappa_M\rangle$, as defined in Equation~\ref{eq:time_series}.
Given this history, a prediction model $\delta_\kappa$ predicts the KPI values for the subsequent $H$ time windows, where $H$ is the forecasting horizon:
\begin{equation}
\langle
\hat{k}^\kappa_{M+1},\ldots,
\hat{k}^\kappa_{M+H}
\rangle
=
\delta_\kappa
\left(
\langle k^\kappa_1,\ldots,k^\kappa_M\rangle
\right),
\end{equation}
%Next, we instantiate this setting for the concurrent-trace and throughput-time KPIs.

\myparit{Concurrent Trace Prediction (CTP)}
This task aims to forecast the number of traces that will be active at future points in time. CTP represents the workload of the process, as all concurrent traces require attention. Accurate predictions can help to forecast overload, if concurrent traces exceed resource capacities, or low utilization, if concurrent traces drop.

We operationalize CTP as a time series forecasting problem by instantiating a process-level KPI $\kappa$ as the number of concurrent traces, denoted by $\kappa_{ct}$. Let $\tau_i$ denote the end of time window $W_i$. A trace is considered active at $\tau_i$ if it has started but not yet completed. The value of $\kappa_{ct}$ for window $W_i$, i.e., the number of active traces at $\tau_i$, is thus defined as
\begin{equation}
ct_i
=
g_{\kappa_{ct}}(\mathcal{L},W_i)
=
\left|
\left\{
\sigma\in\mathcal{L}
\mid
t_s(\sigma)\leq\tau_i<t_c(\sigma)
\right\}
\right|.
\end{equation}

The corresponding concurrent-trace KPI time series is
$K^{\kappa_{ct}}=CT=\langle ct_1,\ldots,ct_M\rangle$.
Given the historical series $CT$, the task is to predict the number of concurrent traces over a forecasting horizon of $H$ days:
$\widehat{CT}
=
\langle
\hat{ct}_{M+1},\ldots,\hat{ct}_{M+H}
\rangle$.

\myparit{Throughput Time Prediction (TTP)}
This task aims to forecast the average completion time of traces in the future. TTP represents process quality in terms of efficiency as higher throughput time indicates extended time requirements. Accurate predictions can help to forecast process-level bottlenecks, e.g., if throughput time is expected to increase.

We operationalize TTP as a time series forecasting problem by instantiating the process-level KPI $\kappa$ as the average throughput time of completed traces, denoted by $\kappa_{tt}$. Let
$\mathcal{C}_i
=
\left\{
\sigma\in\mathcal{L}
\mid
t_c(\sigma)\in W_i
\right\} \neq\emptyset$
denote the set of traces completed during time window $W_i$. The value of $\kappa_{tt}$ for time window $W_i$ is defined as
\begin{equation}
tt_i
=
g_{\kappa_{tt}}(\mathcal{L},W_i)
=
\frac{1}{|\mathcal{C}_i|}
\sum_{\sigma\in\mathcal{C}_i}
d(\sigma).
\end{equation}
%where $d(\sigma)$ is the throughput time of trace $\sigma$.
The corresponding throughput-time KPI time series is
$K^{\kappa_{tt}}=TT=\langle tt_1,\ldots,tt_M\rangle$.
Given the historical series $TT$, the task is to predict the average throughput time over a forecasting horizon of $H$ days:
$\widehat{TT}
=
\langle
\hat{tt}_{M+1},\ldots,\hat{tt}_{M+H}
\rangle$.

\section{Experimental Setup}
\label{sec:eval}
We now present the experimental setup used to compare the ability of different approaches to perform CTP and TTP.
For that, we describe the used data, applied time series splitting, used benchmark approaches, hyperparameter tuning, and finally evaluation metrics. Implementation and experiments can be found in our repository.\footnote{\url{https://github.com/michaelgrohs/process-level-prediction}}
Note that we used \textit{Claude} to assist with the Python implementation; all AI-generated code was reviewed and adapted by the authors.

\mypar{Data}
To assess performance under different levels of control, we experiment with both synthetic and real-life event logs. Synthetic logs are generated from manually constructed process models. This allows us to examine well-defined conditions with control over CTP and TTP time series. 
%This allows us to examine whether the probes can recover process properties under well-defined conditions.
Complementary, we use real-life logs, enabling us to assess performance in realistic and potentially noisier environments. Table \ref{tab:log_stats} shows descriptive statistics of the used logs.

\myparit{Synthetic logs}
To evaluate varying complexity, we generate two synthetic event logs: a \textit{simple} loan application process (\textit{LoAp}) and a more \textit{complex} order-to-cash process (\textit{O2C}). For both, we define a process model to specify the control-flow. 
%LoAp logs are simulated to contain 5,956 traces.
The LoAp model contains ten activities and only one decision to either accept or reject a loan. The O2C model contains 22 activities with multiple decisions, concurrency, and an optional loop. %O2C logs are simulated to contain 19,009 traces. 
We perform a playout of the predefined process models and insert some noise to obtain also less frequent behavior. Time and resource attributes are generated according to predefined distributions. CTP and TTP time series are obtained through aggregation of traces. Thereby, we instantiate both series with time windows $\mathcal{W}$ of calendar days, though other granularities are possible. 
% To obtain logs that have steady time series for CTP and TTP until the last day of the observation period, we clip traces on this day, making the respective KPIs realistic until this day and abruptly stop after. Therefore, the logs also contain traces with only one event because these traces start on the last day of the observation period and are then clipped.
%but the full log useless after this day.
As we aim for time series that remain in a steady state until the end of the observation period, we trim the logs at a fixed cutoff date: all traces are clipped at the last day of the observation period, so that the KPIs are representative until this point and fall to 0 after. 

For both processes, we simulate five scenarios with different trends to see how robust approaches are to changes in process-level behavior:
\begin{inparaenum}
    \item Flat: Similar behavior across all time windows;
    \item Seasonal: Weekly fluctuations in concurrent traces;
    \item Trend: Increasing number of concurrent traces;
    \item Drift: Changing probabilities in decisions in the process;
    \item Combined: All of the above combined.
\end{inparaenum}
Seasonal and trend effects are inflicted to CTP time series primarily, meaning the effects may not be as visible in TTP.
This results in ten logs, five for LoAp and five for O2C.

\myparit{Real-life logs}
%We utilize multiple well-known real-life event logs. In particular, we use the BPIC 2012 with only A\_ activities (bpic12-a)\footnote{\url{https://data.4tu.nl/articles/_/12689204/1}}, the BPIC 2015 municipalities 1 and 2 (bpic15-1 and -2)\footnote{\url{https://data.4tu.nl/collections/_/5065424/1}}, BPIC 2017 Offer log (bpic17-o)\footnote{\url{https://data.4tu.nl/articles/_/12705737/1}}, the BPIC 2020 domestic and international (bpic20-dom and -int)\footnote{\url{https://data.4tu.nl/collections/_/5065541/1}}, the sepsis log\footnote{\url{https://data.4tu.nl/articles/dataset/Sepsis_Cases_-_Event_Log/12707639}} and the helpdesk log\footnote{\url{https://data.4tu.nl/articles/dataset/Dataset_belonging_to_the_help_desk_log_of_an_Italian_Company/12675977}}.
We use multiple real-life logs:\footnote{\url{https://www.tf-pm.org/resources/logs}} BPIC 2012 only A\_ activities (bpi12-a), BPIC 2015 municipalities 1 and 2 (bpi15-1 and -2), BPIC 2017 Offers (bpi17-o), BPIC 2020 domestic and international (bpi20-d and -i), sepsis, and helpdesk.

\begin{table*}[tbp]
\centering
\caption{Statistics of the event logs used in the evaluation. Tr. len. = Trace length; Dur. (d) = Duration in days.}
\label{tab:log_stats}
\resizebox{\textwidth}{!}{%
\setlength{\tabcolsep}{2pt}
\begin{tabular}{@{}clrrrrrrrrrrrrrrrrrrr@{}}
\toprule
& & \multicolumn{10}{c}{\textit{Synthetic}} & & \multicolumn{8}{c}{\textit{Real-life}} \\
\cmidrule(lr){3-12}\cmidrule(lr){14-21}
& & \multicolumn{5}{c}{Loan Application (LoAp)} & \multicolumn{5}{c}{Order-to-Cash (O2C)} &  & \multirow{2}{*}{\rotatebox{90}{bpi12-a\,}} & \multirow{2}{*}{\rotatebox{90}{bpi15-1\,}} & \multirow{2}{*}{\rotatebox{90}{bpi15-2\,}} &  \multirow{2}{*}{\rotatebox{90}{bpi17-o\,}} &  \multirow{2}{*}{\rotatebox{90}{bpi20-d\,}} & \multirow{2}{*}{\rotatebox{90}{bpi20-i\,}} &  \multirow{2}{*}{\rotatebox{90}{helpdesk\,}} & \multirow{2}{*}{\rotatebox{90}{sepsis\,}}  \\
\cmidrule(lr){3-7}\cmidrule(lr){8-12}
\rule{0pt}{1.8em} & & Flat & Seas. & Trend & Drift & Comb. & Flat & Seas. & Trend & Drift & Comb. & &  &  &  &  &  &  &  &  \\
\midrule
\multicolumn{2}{l}{Activities}  & 10 & 10 & 10 & 10 & 10 & 22 & 22 & 22 & 22 & 22 && 10 & 398 & 410 & 8 & 17 & 34 & 14 & 16 \\
\multicolumn{2}{l}{Traces}      & 5{,}956 & 5{,}956 & 5{,}956 & 5{,}956 & 5{,}956 & 19{,}009 & 19{,}009 & 19{,}009 & 19{,}009 & 19{,}009 && 13{,}087 & 1{,}199 & 832 & 31{,}509 & 10{,}500 & 6{,}449 & 4{,}580 & 1{,}050 \\
\multicolumn{2}{l}{Events}      & 46{,}921 & 46{,}799 & 45{,}854 & 44{,}679 & 41{,}800 & 236{,}603 & 235{,}738 & 229{,}684 & 233{,}560 & 211{,}851 && 60{,}849 & 52{,}217 & 44{,}354 & 193{,}849 & 56{,}437 & 72{,}151 & 21{,}348 & 15{,}214 \\
\midrule
\multirow{3}{*}{\rotatebox{90}{Tr.\ len.}}
& Min  & 1 & 2 & 2 & 2 & 1 & 1 & 1 & 1 & 1 & 1 && 3 & 2 & 1 & 3 & 1 & 3 & 2 & 3 \\
& Mean & 7.9 & 7.9 & 7.7 & 7.5 & 7.0 & 12.5 & 12.4 & 12.1 & 12.3 & 11.1 && 4.7 & 43.6 & 53.3 & 6.2 & 5.4 & 11.2 & 4.7 & 14.5 \\
& Max  & 12 & 12 & 12 & 12 & 12 & 28 & 28 & 28 & 29 & 28 && 8 & 101 & 132 & 40 & 24 & 27 & 15 & 185 \\
\midrule
\multirow{3}{*}{\rotatebox{90}{Dur.\ (d)}}
& Min  & 0.0 & 0.1 & 0.1 & 0.2 & 0.0 & 0.0 & 0.0 & 0.0 & 0.0 & 0.0 && 0.0 & 0.4 & 0.0 & 0.0 & 0.0 & 6.7 & 30.6 & 0.0 \\
& Mean & 14.7 & 14.6 & 14.1 & 13.5 & 19.4 & 21.0 & 20.9 & 19.8 & 20.6 & 25.8 && 8.1 & 95.7 & 160.1 & 20.3 & 11.5 & 86.5 & 40.9 & 28.5 \\
& Max  & 68.8 & 68.8 & 68.8 & 68.2 & 93.8 & 118.8 & 118.8 & 118.8 & 140.0 & 136.7 && 91.5 & 1{,}486.0 & 1{,}326.0 & 167.2 & 469.3 & 742.0 & 60.0 & 422.3 \\
\bottomrule
\end{tabular}
}
\end{table*}

\mypar{Splitting}
We split the event logs into a train, validation, and test time window with the first 70\%, the next 10\%, and the last 20\% of the days contained in the log, respectively. Train and validation windows then comprise the historic time series $CT$ and $TT$, the test window comprises $\hat{CT}$ and $\hat{TT}$. For  trace-level approaches, traces are assigned by completion time: the training (validation) set contains all traces that end within the train window (start and end within the validation window), while the test window contains all traces the are active by its start. 

\begin{figure}[bh]
    %\centering
    \begin{subfigure}{.49\textwidth}
        \centering
        \includegraphics[width=\textwidth]{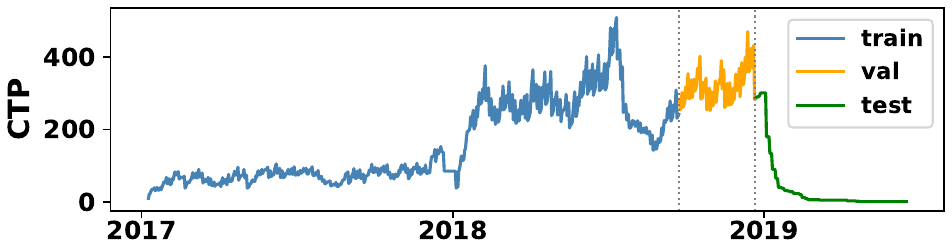}
        \caption{Default splits without truncation.}
        \label{fig:dom-none}
    \end{subfigure}%
    \hfill
    \begin{subfigure}{.49\textwidth}
        \centering
        \includegraphics[width=\textwidth]{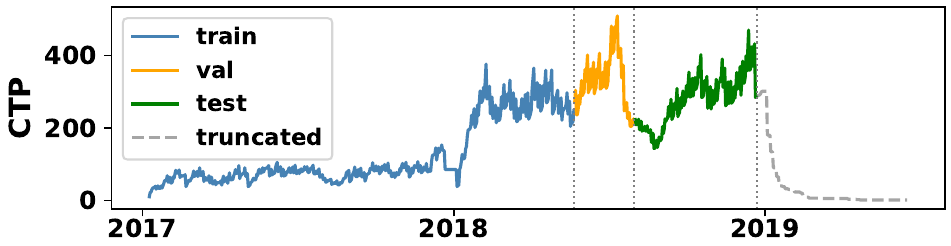}
        \caption{Splits with truncation based on steady-state detection.}
        \label{fig:dom-06}
    \end{subfigure}
    
    \caption{CTP time series in bpi20-d.}
    \label{fig:main}
    
\end{figure}

Splitting may introduce systematic differences in CTP and TTP time series between train, validation, and test window, depending on how the amount of concurrent traces and their average throughput time evolve over the course of the log. This may imply that parts of the time series are not representative and realistic anymore.
In the synthetic logs, we have control over these differences. Here, we explicitly use the five scenarios to test different evolutions. In the real-life logs, however, we encounter a typical issue when handling real-life data: a significant fadeout period due to both stopping of recordings and outliers that are recorded much later than the rest of traces. This means that there may be close to zero active traces in the test period (see Fig.~\ref{fig:dom-none} for bpi20-d), which makes the test window neither meaningful nor realistic.%\footnote{While fadeouts may also affect traditional trace-level PPM, they are usually not accounted for as data is split on trace-level, not day-level as we do here.} %For trace-level PPM  approaches, this effect on test data is not as relevant as metrics are computed per trace and fadeouts do not change the validity of trace-level splits.

To mitigate fadeout effects in real-life logs, we \textit{truncate} the time series and cut off a part of the last windows before splitting. Through that, we ensure that all splits contain behavior from a non-fadeout state. We determine the fadeout periods by using state-of-the-art steady state detection \cite{kraus2026framework}, which classifies the time series into windows that are steady (consistent over time) or unsteady (periods of change). We truncate the last steady period (where the log has faded out already) and the unsteady period before that (the fading out period itself). After that, we split the time series into train, validation, and test set. The result for bpi20-d without a fadeout in the test window is shown in Fig.~\ref{fig:dom-06}.
\footnote{We also used other truncation methods, truncating the time series whenever it falls below a percentage \textit{p} below its peak value or \textit{mag} times its standard deviation below its mean. Results are available in our repository, but since findings are similar to steady state-based truncation, we focus only on this here.}

%We investigate two truncation methods. (1) \textit{peak}: set the last window of the observation period to the first time the CTP time series falls below \textit{p}\% of its peak value, where p is set to 60, 70, and 80. (2) \textit{magnitude}: set the last window of the observation period to the first time the CTP time series falls \textit{mag} times its standard deviation below its mean, where mag is set to 1, 2, and 3. We apply the same truncation also to TTP to ensure the same splits across tasks. Only after the truncation, we split the time series into train, validation, and test set. The result for bpi2020-int when applying peak with X=60 is shown in Fig.~\ref{fig:int-06}, which more closely resembles realistic behavior in all splits.

\mypar{Benchmark approaches}
We investigate six different groups of approaches: baselines, statistical time series models, ML-based time series models, tabular foundation models, process simulation approaches, and trace-level PPM approaches.

\myparit{Na\"{i}ve baselines} We use two naive baselines to establish a lower performance bound.
\begin{compactitem}[$\bullet$]
    \item \texttt{Na\"{i}ve} predicts the last observed value in the validation horizon across the entire test horizon.
    % \item \textit{Seasonal Naive} predicts the same values for every weekday observed in the last observed week of the validation horizon across the entire test horizon.
    \item \texttt{Validation Average} predicts the average of the validation horizon across the entire test horizon.
\end{compactitem}

\myparit{Statistical models}
We include four established statistical models from classical time series literature.
\begin{compactitem}[$\bullet$]
    \item \texttt{Prophet} is a forecasting model designed for time series data with strong seasonal effects and historical trends, decomposing the data into trend, seasonality, holidays, and error~\cite{TaylorL17}.
    \item \texttt{ETS} is an exponential smoothing model that fits errors, trends, and seasonality in additive or multiplicative form \cite{hyndman2002state}.
    \item \texttt{SARIMAX} extends to the widely used ARIMA approach with seasonal and optional exogenous components \cite{alharbi2022seasonal}.
    % \item \textit{Theta} decomposes the series into two theta lines with different curvatures and combines their extrapolations, which is equivalent to simple exponential smoothing with drift~\cite{assimakopoulos2000theta}.
    \item \texttt{STL} decomposes a time series into trend, seasonal, and remainder components for separate forecasting~\cite{cleveland1990stl}.
\end{compactitem}

\myparit{ML models}
We use three classical ML and deep learning models, as used in existing process-level KPI forecasting work.
\begin{compactitem}[$\bullet$]
    \item \texttt{Ridge} applies linear regression recursively (one-step-ahead predictions fed back as input). We also apply a MIMO variant Ridge\textsubscript{M} that predicts the full test horizon jointly in a single forward pass. %, eliminating error accumulation at the cost of a larger output layer.
    % \item \textit{GRU} replaces the linear regression with a gated recurrent network, applied recursively as \textit{Ridge}. We also apply a MIMO variant GRU\textsubscript{M} similar to Ridge\textsubscript{M}.
     \item \texttt{N-BEATS} uses stacks of fully connected residual blocks to produce trend and seasonality components~\cite{oreshkin2019nbeats}.
    %\item \textit{N-HiTS}~\cite{challu2023nhits} is a neural forecasting model based on fully connected residual blocks that combines hierarchical interpolation with multi-rate input sampling to efficiently capture temporal patterns across multiple scales.
    \item \texttt{TFT} is a transformer-based forecasting model that combines recurrent layers, multi-head self-attention, and gating mechanisms to capture both short- and long-term temporal dependencies~\cite{lim2021temporal}. 
\end{compactitem}
% All machine learning-based models are tuned via Bayesian hyperparameter optimisation on the validation set.

\myparit{Tabular foundation models (FM)}
We include two foundation models that operate without dataset-specific training.                                                                          
\begin{compactitem}[$\bullet$]                                       
      \item \texttt{TabPFN}~\cite{Hollmann0EH23} is a tabular in-context learner that performs Bayesian inference over a prior of synthetic datasets via a single transformer forward pass. We use TabPFN-3 and cast the forecasting task as a supervised regression problem using a sliding-window lag representation, fitting one TabPFN estimator per horizon step. 
      
      \item \texttt{Chronos}~\cite{ansari2024chronos} is a time series foundation model that tokenizes series and generates probabilistic forecast trajectories. We use the \texttt{large} variant zero-shot, passing train and validation series as context and reporting the median over 20 trajectories. For test horizons $>$ 64 steps, we apply autoregressive inference, as recommended by the authors.
      
      %\item \texttt{Chronos}~\cite{ansari2024chronos} is a time series foundation model that tokenizes real-valued series and generates probabilistic forecast trajectories. We use the \texttt{large} variant zero-shot, passing the full train and validation series as context and reporting the median over 20 sample trajectories. For test horizons exceeding 64 steps, we apply chunked autoregressive inference, as recommended by the authors. %: each 64-step chunk is predicted conditioned on the training series extended with all previously predicted chunks.
\end{compactitem}

\myparit{Simulation}
% Process simulation can forecast process-level behavior by generating an event log that is the simulated future of a process, learned and fitted on existing event log data. Thus, we can utilize it for CTP and TTP by deriving the corresponding time series from the simulated log in the test period. Rather than treating the tasks as sequences to be extrapolated, simulation approaches encode explicit relationships between process features like arrivals, departures, and work-in-progress to project them forward jointly. 
%This may make simulation robust for long-term process-level predictions. 
We include three simulation approaches based on different paradigms.
\begin{compactitem}[$\bullet$]
     \item \texttt{PMSD}~\cite{pourbafrani2022discovering} discovers a system dynamics model and simulates the process by fitting a set of regression equations between variables like active resources, throughput time, finish and arrival rates. %Since the original approach requires a person to manually turn these relations into a simulation diagram and run it in dedicated third-party simulation software, which does not scale to running one benchmark unattended across many datasets, we instead fix a default structure across all datasets and evaluate it ourselves with a simple recursive loop, rather than a full case-level simulation of the process.
     We obtain CTP and TTP time series through the fitted model: concurrent traces through finish and arrival rates, throughput time directly.
    \item \texttt{SIMOD}~\cite{Simod} discovers a BPMN model and a set of simulation parameters from the training log and generates a simulated event log that represents process executions during the test period.
    \item \texttt{ASIM}~\cite{KirchdorferBKAS24} discovers a multi-agent system, where each agent represents a process resource. By means of agent actions and interactions, the approach simulates the future process evolution, resulting in a simulated event log.\footnote{Note that we apply ASIM only to synthetic logs due to missing event start timestamps in real-life logs.}
\end{compactitem}
For Simod and ASIM, we derive CTP and TTP time series from the simulated log whereas PMSD provides them directly.

\myparit{Trace-level PPM approaches}
Typical trace-level PPM approaches are trained on prefixes generated from completed traces, i.e., incomplete versions of the traces. Based on these prefixes, a model is trained in a supervised manner to predict the future of the prefixes.
%Formally, for a given trace $\sigma$, its event prefix of length $m \in [1, n-1]$ is defined as $\sigma^m=\langle e_{1}, \ldots, e_{m}\rangle$, representing a partial execution up to the $m$-th event. 
% Each event prefix is then encoded into a feature vector $\Gamma(\sigma^m) = \bm{x} \in \mathcal{X}$ and associated with a target value $y \in \mathcal{Y}$. 
Here, we use two common PPM tasks that can be aggregated into CTP and TTP:
\begin{compactenum}[(1)]
    \item \textit{Remaining time prediction} estimates the time between the current event and the completion time of a trace.
    %function $\delta_{rt}$ to estimate the time interval between the current event and the last event of the trace: %$\hat{y}_{rt} = 
    %$\delta_{rt}(\sigma^m) \approx \pi_T(e_{n}) - \pi_T(e_m)$. 
    \item \textit{Suffix prediction} predicts the complete event suffix from the current event until the trace's final event.
    %function $\delta_{su}$ to predict the complete event suffix from the current event until the trace's final event: %$\hat{y}_{su} = 
    %$\delta_{su}(\sigma^m) \approx \langle e_{m+1}, ..., e_{n}\rangle$.
\end{compactenum}

\noindent
We use three different trace-level PPM approaches and then aggregate the trace-level predictions into CTP and TTP.
\begin{compactitem}[$\bullet$]
    \item \texttt{GLSTM} is based on a generative LSTM~\cite{camargo_DGEN}, autoregressively predicting the suffix of a trace.
    \item \texttt{PT} employs a transformer~\cite{bukhsh2021processtransformer}, which we use for both suffix (\textbf{PT\textsubscript{su}}) and remaining time prediction (\textbf{PT\textsubscript{RT}}).
    \item \texttt{PGT} is a state-of-the-art remaining time prediction approach based on a graph transformer architecture \cite{PGTNet}.
\end{compactitem}

\noindent
These trace-level approaches are designed to predict the future of running traces, so they expect non-empty prefixes. 
To account for traces that begin in the test horizon, we test three different configurations: 
%for providing prefixes of traces that start during the test window:
\begin{compactitem}[$\bullet$]
    \item \textit{Plain}: Following recent works~\cite{KirchdorferOAS26}, we predict the daily arrival rate in the test window with Prophet~\cite{TaylorL17}  and
    %which are distributed across each day according to the empirical intraday arrival distribution observed in the training data. 
    instantiate as many one-event prefixes per day as predicted to arrive. 
    The activity is set to the most frequent start activity, the resource is sampled from the activity's empirical distribution.
    \item \textit{First}: We provide the actual first event of each trace.
    \item \textit{Half}: We provide the actual first half of each trace.
\end{compactitem}

\noindent
\textit{Plain} relies only on predicted start events; \textit{First} and \textit{Half} provide ground-truth information from the test window and thus represent oracle-assisted settings that give the PPM approaches an informational advantage. In other words, \textit{Plain} reflects a realistic deployment scenario, whereas the oracle-assisted settings assess the upper bound of trace-level PPM.

%To aggregate trace-level predictions into process-level KPIs, we project the remaining time prediction or the last timestamp of a predicted suffix on the days the trace is predicted active and the throughput time on its predicted end day, thus obtaining CTP and TTP predictions on an aggregated level, respectively.
To aggregate trace-level predictions into process-level KPIs, we use each trace's predicted active period and completion time. For CTP, we mark a trace as active on each day covered by its predicted remaining time (or, equivalently, up to the last timestamp of its predicted suffix). For TTP, we assign the trace's predicted throughput time to its predicted end day.

\mypar{Hyperparameter tuning}
We tune hyperparameters on the validation set for most approaches. For statistical and ML models, we use grid search over model size and lookback and forecast window length. For TabPFN, we tune the lookback window. For the PPM approaches, we tune model size and learning rate. Details are provided in our repository.

\mypar{Metrics}
We evaluate prediction accuracy using Mean Absolute Error (MAE). %\footnote{Additional metrics like Mean Squared Error can be found in our repository.}.
For CTP, MAE is expressed in units of traces (i.e., how many active traces the prediction deviates from the ground truth on average per day). For TTP, MAE is expressed in hours (i.e., deviation between predicted and actual daily mean trace duration).
To additionally assess whether a model systematically over- or under-predicts, we report the Normalized Mean Bias Error (NMBE).
Unlike MAE, NMBE preserves the direction of errors rather than taking their absolute value: a positive NMBE indicates systematic under-prediction, while a negative NMBE indicates systematic over-prediction. %NMBE therefore complements MAE rather than replacing it, as it can approach zero even for models with substantial errors, provided that over- and under-predictions balance out.

\section{Results}
\label{sec:results}
In this section, we present the experimental results by discussing prediction accuracy and error direction.
%We report results from one random seed. Our repository shows the results from two more seeds, indicating robustness to random weight initialization.

% ===== Concurrent Cases =====
\begin{table*}[b]
\centering
\caption{Evaluation results for \textit{concurrent traces} across synthetic event logs. \textbf{Bold}: best per column; \underline{underline}: second-best.}
\label{tab:results_cc}
\resizebox{\textwidth}{!}{%
\setlength{\tabcolsep}{4pt}
\begin{tabular}{@{}cclrrrrrr|rrrrrrrr@{}}
\toprule
& & & \multicolumn{5}{c}{\textit{LoAp (simple)}} & \multicolumn{1}{c}{\multirow{2}{*}{Avg.}} & \multicolumn{5}{c}{\textit{O2C (complex)}} & \multirow{2}{*}{Avg.} & \\
\cmidrule(lr){4-8}\cmidrule(lr){10-14}
& & & Flat & Season. & Trend & Drift & Comb. & \multicolumn{1}{c}{} & Flat & Season. & Trend & Drift & Comb. &  \\
\midrule
\multicolumn{2}{c}{\multirow{2}{*}{\rotatebox[origin=c]{90}{Basel.}}} & Na\"{i}ve
& 14.13 & 15.38 & 45.93 & 15.71 & 187.47 & 55.72
& 26.43 & 80.89 & 485.04 & 48.53 & 1{,}021.45 & 332.47 \\
& & Val. Avg.
& 27.71 & 29.65 & 109.32 & 23.99 & 254.37 & 89.01
& \underline{22.55} & 48.64 & 596.61 & 47.06 & 1{,}337.14 & 410.40 \\
\midrule
\multicolumn{2}{c}{\multirow{4}{*}{\rotatebox[origin=c]{90}{Statistical}}}
& Prophet
& 645.10 & 561.89 & 121.06 & 173.59 & 23.37 & 305.00
& 1{,}498.57 & 1{,}536.06 & 1{,}747.54 & 395.31 & 637.37 & 1{,}162.97 \\
& & ETS
& 14.13 & \textbf{10.97} & 345.35 & 49.73 & 247.01 & 133.44
& \textbf{21.31} & \underline{40.69} & 171.14 & 32.94 & 172.44 & 87.70 \\
& & SARIMAX
& 75.81 & 101.08 & 44.94 & 99.24 & \textbf{12.36} & 66.69
& 73.04 & 41.21 & \textbf{71.62} & 67.69 & 105.79 & \underline{71.87} \\
& & STL 
& 14.12 & 13.75 & 48.28 & \underline{12.96} & 45.30 & \underline{26.88} 
& 24.76 & \textbf{22.25} & 159.55 & \underline{32.59} & \underline{100.02} & \textbf{67.84} \\
\midrule
\multicolumn{2}{c}{\multirow{4}{*}{\rotatebox[origin=c]{90}{ML}}}
& Ridge 
& \underline{12.43} & 15.26 & \textbf{17.68} & 23.99 & 85.28 
& 30.93 & 67.84 & 124.77 & \underline{114.98} & 128.07 & 236.02 & 134.34 \\
& & Ridge$_\text{M}$
& 18.50 & 21.87 & \underline{19.62} & 23.49 & 96.44 & 35.98
& 53.89 & 54.68 & 1{,}100.86 & \textbf{22.85} & 1{,}334.02 & 513.26 \\
& & N-BEATS
& 18.50 & 18.66 & 42.83 & 22.26 & \underline{19.80} & \textbf{24.41}
& 33.54 & 63.15 & 956.25 & 42.03 & 354.69 & 289.93 \\
& & TFT
& 31.03 & 32.31 & 100.42 & 51.66 & 232.00 & 89.49
& 47.50 & 72.03 & 545.08 & 62.58 & 1{,}165.24 & 378.49 \\
\midrule
\multicolumn{2}{c}{\multirow{2}{*}{\rotatebox[origin=c]{90}{FM}}}
& Chronos
& \textbf{11.98} & 13.02 & 57.51 & \textbf{11.39} & 67.01 & 32.18
& 49.63 & 146.87 & 506.28 & 65.80 & 277.11 & 209.14 \\
& & TabPFN
& 21.27 & 19.48 & 138.33 & 34.19 & 168.98 & 76.45
& 69.61 & 66.14 & 1{,}357.05 & 65.60 & 1{,}134.64 & 538.61 \\
\midrule
\multicolumn{2}{c}{\multirow{3}{*}{\rotatebox[origin=c]{90}{SIM}}}
& PMSD
& 19.75 & \underline{12.83} & 62.91 & 22.08 & 79.70 & 39.45
& 47.24 & 70.16 & 263.24 & 32.72 & \textbf{70.97} & 96.87 \\
& & Simod
& 680.33 & 632.28 & 244.40 & 615.15 & 419.55 & 518.34
& 993.10 & 1{,}415.46 & 2{,}090.27 & 1{,}096.04 & 3{,}618.08 & 1{,}842.59 \\
& & ASIM
& 150.35 & 100.89 & 181.67 & 236.58 & 515.89 & 237.08
& 744.40 & 963.51 & 701.74 & 1{,}347.46 & 2{,}400.08 & 1{,}231.44 \\
\midrule
\multirow{12}{*}{\rotatebox[origin=c]{90}{Trace-Level PPM}}
& \multirow{4}{*}{\rotatebox[origin=c]{90}{Plain}}
& GLSTM
& 204.33 & 243.80 & 222.49 & 172.80 & 634.60 & 295.48
& 1{,}000.53 & 1{,}661.30 & 2{,}363.32 & 1{,}442.43 & 3{,}783.91 & 2{,}050.30 \\
& & PGT
& 131.44 & 89.89 & 168.04 & 87.49 & 359.36 & 167.24
& 470.77 & 1{,}058.98 & 1{,}939.38 & 1{,}033.06 & 3{,}199.43 & 1{,}540.32 \\
& & PT$_\text{su}$
& 96.91 & 104.98 & 30.04 & 83.38 & 295.33 & 122.13
& 382.91 & 1{,}153.23 & 2{,}006.96 & 1{,}106.49 & 2{,}702.57 & 1{,}470.43 \\
& & PT$_\text{RT}$
& 84.71 & 168.49 & 111.44 & 150.69 & 398.98 & 182.86
& 270.53 & 1{,}564.83 & 2{,}330.85 & 1{,}420.15 & 3{,}143.62 & 1{,}746.00 \\
\cmidrule(l){2-15}
& \multirow{4}{*}{\rotatebox[origin=c]{90}{First}}
& GLSTM
& 262.73 & 253.58 & 315.00 & 179.27 & 673.20 & 336.76
& 1{,}175.77 & 1{,}407.98 & 1{,}690.26 & 1{,}107.83 & 3{,}038.87 & 1{,}684.14 \\
& & PGT
& 60.91 & 61.09 & 92.13 & 37.38 & 428.20 & 135.94
& 409.91 & 387.53 & 413.60 & 374.17 & 1{,}729.13 & 662.87 \\
& & PT$_\text{su}$
& 36.38 & 118.53 & 129.04 & 96.20 & 359.29 & 147.89
& 587.87 & 763.38 & 1{,}254.09 & 752.36 & 1{,}570.72 & 985.69 \\
& & PT$_\text{RT}$
& 94.42 & 181.49 & 107.44 & 163.38 & 462.49 & 201.84
& 413.62 & 1{,}241.11 & 1{,}579.23 & 1{,}032.74 & 1{,}848.02 & 1{,}222.94 \\
\cmidrule(l){2-15}
& \multirow{4}{*}{\rotatebox[origin=c]{90}{\shortstack{Half}}}
& GLSTM
& 100.40 & 114.24 & 149.24 & 94.67 & 365.91 & 164.89
& 709.11 & 691.32 & 656.87 & 638.28 & 2{,}215.28 & 982.17 \\
& & PGT
& 111.00 & 113.24 & 223.27 & 99.76 & 350.42 & 179.54
& 450.23 & 397.11 & 590.40 & 398.30 & 1{,}303.89 & 627.99 \\
& & PT$_\text{su}$
& 164.04 & 96.80 & 164.27 & 111.58 & 125.73 & 132.48
& 87.53 & 204.81 & 418.38 & 154.89 & 676.09 & 308.34 \\
& & PT$_\text{RT}$
& 72.69 & 122.80 & 99.67 & 93.84 & 279.89 & 133.78
& 140.87 & 824.77 & 1{,}055.11 & 693.26 & 1{,}060.34 & 754.87 \\
\bottomrule
\end{tabular}
}
\end{table*}

% ===== Throughput Time =====
\begin{table*}[b]
\centering
\caption{Evaluation results for \textit{avg. throughput time (h)} across synthetic event logs. \textbf{Bold}: best per column; \underline{underline}: second-best.}
\label{tab:results_tt}
\resizebox{\textwidth}{!}{%
\setlength{\tabcolsep}{4pt}
\begin{tabular}{@{}cclrrrrrr|rrrrrrrr@{}}
\toprule
& & & \multicolumn{5}{c}{\textit{LoAp (simple)}} & \multicolumn{1}{c}{\multirow{2}{*}{Avg.}} & \multicolumn{5}{c}{\textit{O2C (complex)}} & \multirow{2}{*}{Avg.} & \\
\cmidrule(lr){4-8}\cmidrule(lr){10-14}
& & & Flat & Seasonal & Trend & Drift & Combined & \multicolumn{1}{c}{} & Flat & Seasonal & Trend & Drift & Combined &  \\
\midrule
\multicolumn{2}{c}{\multirow{2}{*}{\rotatebox[origin=c]{90}{Basel.}}} & Na\"{i}ve
& 92.34 & 58.04 & 33.24 & 63.00 & 86.65 & 66.65
& 81.64 & \underline{42.29} & 51.70 & 47.94 & 101.62 & 65.04 \\
& & Val. Avg.
& \underline{43.00} & \underline{40.74} & 33.82 & \underline{47.40} & 94.14 & 51.82
& \underline{36.77} & 42.62 & \underline{50.37} & 43.04 & 85.91 & \underline{51.74} \\
\midrule
\multicolumn{2}{c}{\multirow{4}{*}{\rotatebox[origin=c]{90}{Statistical}}}
& Prophet
& 56.33 & 67.51 & 229.77 & 132.85 & 89.45 & 115.18
& 94.73 & 156.09 & 180.29 & 190.34 & 199.96 & 164.28 \\
& & ETS
& 46.19 & 51.99 & 33.11 & 47.52 & 83.13 & 52.39
& \textbf{36.38} & 43.16 & \textbf{43.26} & \underline{42.25} & 151.39 & 63.29 \\
& & SARIMAX
& 53.14 & 63.40 & 42.29 & 47.42 & \textbf{73.90} & 56.02
& 139.97 & 109.10 & 148.83 & 52.36 & \underline{78.10} & 105.67 \\
& & STL
& 54.75 & 54.04 & \textbf{30.59} & 111.69 & 124.44 & 75.10
& 48.74 & 47.45 & 55.05 & 52.77 & 80.50 & 56.90 \\
\midrule
\multicolumn{2}{c}{\multirow{4}{*}{\rotatebox[origin=c]{90}{ML}}}
& Ridge
& 51.73 & \textbf{40.73} & \underline{31.99} & 47.68 & 86.28 & \underline{51.68}
& 37.86 & 55.91 & 52.74 & 62.22 & 114.88 & 64.72 \\
& & Ridge$_\text{M}$
& 48.86 & 42.56 & 36.24 & 51.00 & 144.09 & 64.55
& 43.51 & 44.61 & 52.31 & 44.88 & 107.14 & 58.49 \\
& & N-BEATS
& \textbf{42.31} & 43.32 & 38.71 & 48.71 & 114.66 & 57.54
& 38.17 & 46.12 & 50.99 & 43.16 & 105.94 & 56.87 \\
& & TFT
& 52.15 & 57.90 & 51.41 & 65.67 & 134.85 & 72.30
& 47.62 & 54.09 & 68.52 & 54.04 & 125.42 & 69.94 \\
\midrule
\multicolumn{2}{c}{\multirow{2}{*}{\rotatebox[origin=c]{90}{FM}}}
& Chronos
& 48.43 & 47.10 & 42.27 & 47.97 & \underline{74.17} & 51.99
& 50.84 & 44.56 & 59.37 & \textbf{40.93} & \textbf{62.50} & \textbf{51.64} \\
& & TabPFN
& 44.74 & 43.60 & 35.63 & 50.60 & 166.48 & 68.21
& 37.22 & 44.05 & 50.64 & 42.61 & 165.18 & 67.94 \\
\midrule
\multicolumn{2}{c}{\multirow{3}{*}{\rotatebox[origin=c]{90}{SIM}}}
& PMSD
& 44.29 & 47.45 & 45.04 & \textbf{47.26} & \underline{71.49} & \textbf{51.11}
& 42.41 & \textbf{41.28} & 182.51 & 47.06 & 213.28 & 105.31 \\
& & Simod
& 384.27 & 1{,}681.11 & 130.97 & 1{,}201.18 & 503.92 & 780.29
& 724.51 & 551.85 & 409.71 & 426.32 & 307.48 & 483.97 \\
& & ASIM
& 110.82 & 136.71 & 107.66 & 95.43 & 337.59 & 157.64
& 297.86 & 299.26 & 277.05 & 303.15 & 541.09 & 343.68 \\
\midrule
\multirow{12}{*}{\rotatebox[origin=c]{90}{Trace-Level PPM}}
& \multirow{4}{*}{\rotatebox[origin=c]{90}{Plain}}
& GLSTM
& 237.63 & 233.78 & 180.20 & 161.39 & 376.73 & 237.95
& 303.00 & 366.23 & 294.19 & 273.61 & 473.86 & 342.18 \\
& & PGT
& 64.14 & 61.42 & 42.75 & 55.25 & 209.01 & 86.51
& 76.76 & 72.64 & 58.17 & 65.66 & 196.56 & 93.96 \\
& & PT$_\text{su}$
& 54.11 & 109.91 & 79.41 & 100.21 & 169.72 & 102.67
& 140.94 & 240.59 & 222.93 & 292.05 & 212.19 & 221.74 \\
& & PT$_\text{RT}$
& 82.32 & 168.76 & 58.80 & 144.66 & 233.52 & 137.61
& 84.65 & 332.39 & 287.45 & 271.94 & 228.25 & 240.94 \\
\cmidrule(l){2-15}
& \multirow{4}{*}{\rotatebox[origin=c]{90}{First}}
& GLSTM
& 176.18 & 194.28 & 159.51 & 149.34 & 325.87 & 201.04
& 301.91 & 315.44 & 225.53 & 284.70 & 481.08 & 321.73 \\
& & PGT
& 64.14 & 61.42 & 42.75 & 55.25 & 209.01 & 86.51
& 76.76 & 71.19 & 58.25 & 65.58 & 194.19 & 93.19 \\
& & PT$_\text{su}$
& 52.85 & 173.27 & 155.43 & 125.06 & 270.70 & 155.46
& 125.12 & 310.09 & 311.42 & 310.33 & 246.05 & 260.60 \\
& & PT$_\text{RT}$
& 82.32 & 168.76 & 59.45 & 144.66 & 233.52 & 137.74
& 84.23 & 333.59 & 289.65 & 273.33 & 230.04 & 242.17 \\
\cmidrule(l){2-15}
& \multirow{4}{*}{\rotatebox[origin=c]{90}{\shortstack{Half}}}
& GLSTM
& 97.34 & 107.73 & 84.86 & 85.08 & 187.51 & 112.50
& 173.98 & 169.15 & 104.38 & 150.38 & 315.72 & 182.72 \\
& & PGT
& 102.86 & 106.63 & 122.95 & 87.01 & 176.66 & 119.22
& 95.80 & 86.44 & 87.83 & 81.19 & 123.63 & 94.98 \\
& & PT$_\text{su}$
& 120.88 & 78.93 & 80.86 & 98.48 & 91.18 & 94.07
& 82.37 & 101.36 & 119.32 & 81.93 & 118.15 & 100.63 \\
& & PT$_\text{RT}$
& 73.95 & 111.69 & 58.06 & 80.01 & 137.29 & 92.20
& 44.76 & 199.12 & 176.01 & 163.27 & 95.97 & 135.83 \\
\bottomrule
\end{tabular}
}
\end{table*}

\mypar{Prediction accuracy} We first discuss accuracy in synthetic and real-life logs. 

\myparit{Synthetic logs} 
Tab. \ref{tab:results_cc} and \ref{tab:results_tt} report the MAE of all approaches for CTP and TTP in the synthetic logs, respectively.
For CTP, statistical and ML methods, as well as Chronos and PMSD, perform best across most logs, whereas trace-level PPM performs substantially worse.
N-BEATS achieves the lowest average MAE on LoAp (24.41), while STL performs best on O2C (67.84).
Among simulation models, only PMSD is competitive, likely since it directly operates on process-level. In contrast, SIMOD and ASIM exhibit considerably higher MAEs, arising from small errors on trace-level that compound on process-level. 
The same effect leads to high errors in all trace-level PPM configurations, which are often outperformed by the baselines. 
%Their CTP forecasts show substantial errors across configurations and settings. 
Even the \textit{Half} configuration shows little improvement, as models tend to underestimate how long half-completed traces remain active, suggesting a systematic bias toward shorter remaining times.

Trend and combined scenarios show higher MAEs, especially for the baselines, Ridge\textsubscript{M}, TFT, and trace-level PPM approaches. Some approaches like ETS also perform worse in few scenarios (e.g., an MAE of 345.35 on LoAp-trend), decreasing its overall average despite often performing best or second best. This also holds for the ML approaches, which often are among the top two approaches. Prophet is the worst time series approaches, often performing worse than the baselines, likely since it mistakes the ramp-up period as a trend. Chronos and TabPFN are competitive in many logs, but also have few outliers with high MAEs in O2C trend and combined. Overall, statistical, ML, and FM approaches typically outperform the baselines, indicating that they capture meaningful signals and not trivial patterns.

For TTP, these findings largely persist with one major distinction: Val.\ Avg.\ achieves a relatively low average error. In other words, a trivial baseline can compete against more sophisticated approaches. This is because the seasonal, trend, and drift effects impact CTP more than TTP, meaning that the TTP series are relatively flat and thus easy to forecast with the validation average. Notably, approaches do not seem to capture this flat evolution of TTP sufficiently.
This finding may not hold in more complex scenarios, such as the real-life logs. Val. Avg. also never performs best for a specific log. Rather, other approaches work best, with ETS, SARIMAX, Ridge, Chronos, and PMSD standing out. The PPM approaches trail again, but not as substantially as for CTP. On LoAp-flat, PT\textsubscript{su}'s MAE with \textit{Plain} (54.11) comes close to the best MAE of 42.31. In O2C, differences are higher. Performance improves with more observed prefix---GLSTM's average drops from 238 in LoAp and 342 in O2C (\textit{Plain}) to 113 and 183 (\textit{Half}), respectively---but even \textit{Half} remains uncompetitive with direct process-level forecasting. Interestingly, \textit{Half} for PGT performs worse than \textit{First} and \textit{Plain}: it predicts the majority of \textit{Half} traces to finish quickly, while not predicting traces with longer remaining time.

\begin{figure}[b]
    %\centering
    \begin{subfigure}{\textwidth}
        \centering
        \includegraphics[width=\textwidth]{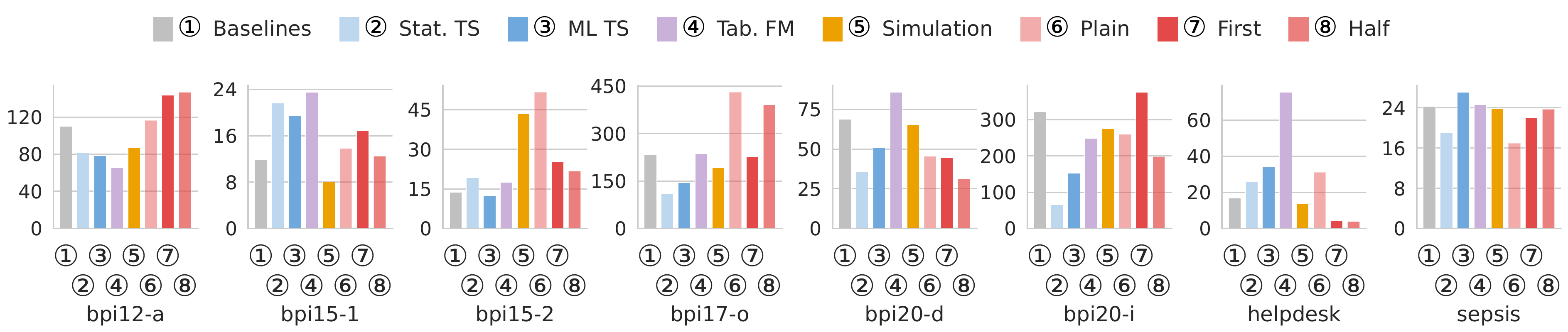}
        \caption{MAE for CTP across all real-life logs.}
        \label{fig:results_real-ctp-ssd}
    \end{subfigure}%
    \hfill
    \begin{subfigure}{\textwidth}
        \centering
        \includegraphics[width=\textwidth]{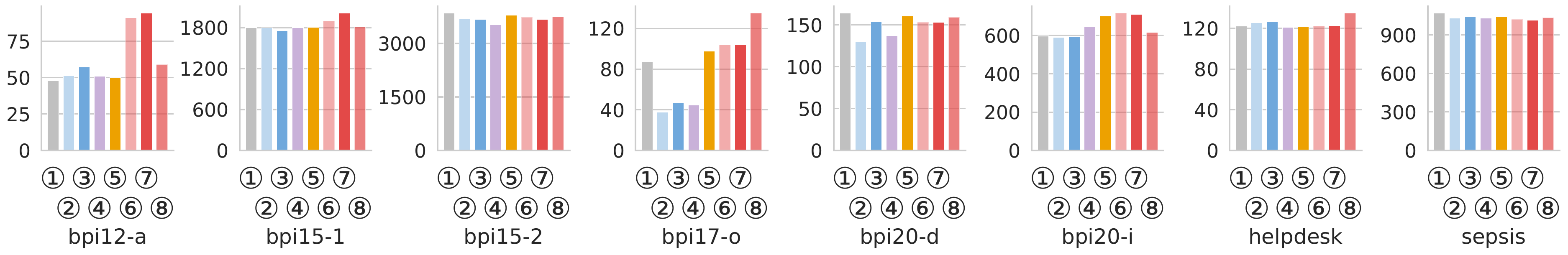}
        \caption{MAE for TTP across all real-life logs.}
        \label{fig:results_real-ttp-ssd}
    \end{subfigure}

    \caption{MAE across all real-life logs; shown is the best MAE per approach category.}
    \label{fig:results_real-ssd}
\end{figure}

\myparit{Real-life logs} Fig. \ref{fig:results_real-ssd} shows the average MAE of the best performing approach per group in the real-life logs.
For CTP, trace-level PPM is not competitive in five logs (bpi12-a, bpi15-1 and -2, bpi17-o, and bpi20-i). Three logs reverse this pattern. First, on helpdesk, PGT \textit{Half} (MAE of 3.9), PGT \textit{First} (4.3), and PT\textsubscript{RT} \textit{First} outperform the best non-PPM approach, PMSD (13.6). This is due to a log characteristic: the trace duration is evenly distributed, meaning trace-level errors cancel out on process-level if the mean trace duration is predicted, which the approaches do. 
Second, on bpi20-d, PGT \textit{Half} (31.3) and PT\textsubscript{RT} (36.5) beat the next best SARIMAX (40.83). This is again due to a log characteristic: there are evenly many short and long traces, and predicting the mean for both means that errors cancel out on process-level. Another factor is a strong increase in the arrival rate in the test period, which \textit{Half} and \textit{First} know per definition.
Third, on sepsis, PGT \textit{Plain} (16.93) beats Ridge (19.59) due to a bi-directional reason: although PGT underpredicts how long traces are active, the forecasted arrival rate is too high, which cancels out on process-level and leads to accurate CTP predictions.             
Advantages in these three logs are not as clear in TTP. On helpdesk, PGT trails ETS; on bpi20-d, Chronos and Prohpet beat PGT; only on sepsis PGT keeps a (narrow, 2\%) lead. This is since TTP is more erratic: CTP is a slowly-changing stock variable with strong day-to-day persistence, which both eases forecasting and enables trace-level cancellation effects, whereas TTP creates daily averages of completing traces, so one trace-level error is more visible.

The other approaches are not affected by trace-level but rather process-level log characteristics. For example, statistical, ML, and FM models work well on bpi12-a and bpi17-o, which are logs with little shifts between train and test, as well as low volatility.
Further, the baselines perform worse for TTP. Only for bpi12-a, Val. Avg. show the lowest MAE, minimally better than PMSD's. In all other logs, other models perform better. This indicates that real-life scenarios require more sophisticated methods. Also, PMSD shows a relatively high MAE for CTP in bpi20-i, bpi15-2, and sepsis. 

Overall, we see that the trace-level approaches again perform worse than process-level models. They are only competitive if data characteristics favor their trace-level idea, which does not generalize across all logs.

\mypar{Error direction} We first analyze the direction of process-level errors and then trace-level PPM errors.

\myparit{Process-level errors} \autoref{tab:results_nmbe} reports the median NMBE across all used logs, indicating whether approaches systematically over- or underestimate CTP and TTP.
In the synthetic logs, baselines, statistical models, ML models, Chronos, TabPFN, and PMSD all center close to zero (median NMBE within $\pm0.05$), indicating no consistent bias. Prophet is the exception, overestimating substantially (median NMBE $+0.32$ for CTP). The trace-level PPM approaches, in contrast, systematically underestimate CTP in nearly every configuration, with the bias shrinking as more prefix is observed, e.g., GLSTM improves from $-0.55$ (\textit{Plain}) to $-0.29$ (\textit{Half}) and PT\textsubscript{su} becomes essentially unbiased ($+0.05$) in \textit{Half}. On the real-life logs, this pattern persists: the PPM approaches underestimate CTP (median NMBE $-0.55$ to $-0.05$), whereas the other approaches shift to (mild) \emph{over}estimation (median NMBE $+0.03$ to $+0.44$), with PMSD ($+0.01$) as the only near-unbiased method. For TTP, the process-level approaches remain mostly near zero on both synthetic and real-life logs, whereas the PPM approaches underestimate (median NMBE $-0.56$ to $-0.03$ on synthetic, $-0.54$ to $-0.10$ on real-life). Interestingly, PGT stands out for TTP with a median NMBE close to 0 for \textit{First} and \textit{Plain}, indicating no systematic bias on average.

\begin{table*}[!bt]
\centering
\caption{Median NMBE across synthetic and real-life event logs. Positive values indicate systematic overestimation ($\geq0.10$ highlighted in blue), negative values underestimation ($\leq-0.10$ highlighted in red).}
\label{tab:results_nmbe}
\resizebox{\textwidth}{!}{%
\setlength{\tabcolsep}{3pt}
\begin{tabular}{@{}cllrrrrrrrrrrrrrrrrrrrrrrrrrrr@{}}
\toprule
& & & \multicolumn{2}{c}{Basel.} & \multicolumn{4}{c}{Statistical} & \multicolumn{4}{c}{ML} & \multicolumn{2}{c}{FM} & \multicolumn{3}{c}{SIM} & \multicolumn{12}{c}{Trace-Level PPM} \\
\cmidrule(lr){4-5}\cmidrule(lr){6-9}\cmidrule(lr){10-13}\cmidrule(lr){14-15}\cmidrule(lr){16-18}\cmidrule(lr){19-30}
& & & & & & & & & & & & & & & & & & \multicolumn{4}{c}{Plain} & \multicolumn{4}{c}{First} & \multicolumn{4}{c}{Half} \\
\cmidrule(lr){19-22}\cmidrule(lr){23-26}\cmidrule(lr){27-30}
\rule{0pt}{0em}
& & 
& \rotatebox{90}{Na\"{i}ve\,}
& \rotatebox{90}{Val.\ Avg.\,}
& \rotatebox{90}{Prophet\,}
& \rotatebox{90}{ETS\,}
& \rotatebox{90}{SARIMAX\,}
& \rotatebox{90}{STL\,}
& \rotatebox{90}{Ridge\,}
& \rotatebox{90}{Ridge$_\text{M}$\,}
& \rotatebox{90}{N-BEATS\,}
& \rotatebox{90}{TFT\,}
& \rotatebox{90}{Chronos\,}
& \rotatebox{90}{TabPFN\,}
& \rotatebox{90}{PMSD\,}
& \rotatebox{90}{Simod\,}
& \rotatebox{90}{ASIM\,}
& \rotatebox{90}{GLSTM\,}
& \rotatebox{90}{PGT\,}
& \rotatebox{90}{PT$_\text{RT}$\,}
& \rotatebox{90}{PT$_\text{su}$\,}
& \rotatebox{90}{GLSTM\,}
& \rotatebox{90}{PGT\,}
& \rotatebox{90}{PT$_\text{su}$\,}
& \rotatebox{90}{PT$_\text{RT}$\,}
& \rotatebox{90}{GLSTM\,}
& \rotatebox{90}{PGT\,}
& \rotatebox{90}{PT$_\text{su}$\,}
& \rotatebox{90}{PT$_\text{RT}$\,} \\
\midrule
\multirow{2}{*}{\rotatebox[origin=c]{90}{Syn.}}
& & CTP
& $-$.03 & $-$.01 & \pos{.32} & $-$.01 & $-$.02 & .00
& .03 & .01 & .02 & $-$.02
& $-$.03 & $-$.03
& .00 & \nega{$-$.45} & \nega{$-$.14}
& \nega{$-$.55} & \nega{$-$.22} & \nega{$-$.39} & \nega{$-$.26}
& \nega{$-$.52} & \nega{$-$.17} & \nega{$-$.32} & \nega{$-$.43}
& \nega{$-$.29} & \nega{$-$.25} & .05 & \nega{$-$.26} \\
& & TTP
& .01 & $-$.03 & \pos{.28} & .01 & $-$.04 & .03
& .01 & .01 & $-$.03 & $-$.01
& $-$.06 & $-$.01
& $-$.01 & \pos{.83} & \nega{$-$.55}
& \nega{$-$.56} & \nega{$-$.11} & \nega{$-$.42} & $-$.09
& \nega{$-$.53} & \nega{$-$.11} & $-$.03 & \nega{$-$.42}
& \nega{$-$.28} & \nega{$-$.22} & $-$.03 & \nega{$-$.23} \\
\midrule
\multirow{2}{*}{\rotatebox[origin=c]{90}{Real}}
& & CTP
& \pos{.23} & \pos{.16} & \pos{.44} & \pos{.29} & \pos{.25} & .03
& \pos{.24} & \pos{.21} & \pos{.27} & \pos{.24}
& \pos{.20} & \pos{.31}
& .01 & \pos{.16} & --
& \nega{$-$.51} & $-$.05 & \nega{$-$.39} & \nega{$-$.32}
& \nega{$-$.55} & $-$.09 & \nega{$-$.38} & \nega{$-$.34}
& \nega{$-$.29} & \nega{$-$.16} & \pos{.19} & \nega{$-$.14} \\
& & TTP
& \nega{$-$.15} & $-$.06 & \pos{.11} & $-$.07 & \nega{$-$.21} & $-$.08
& \nega{$-$.11} & $-$.09 & \nega{$-$.18} & \nega{$-$.15}
& \nega{$-$.18} & $-$.07
& $-$.08 & \nega{-.21} & --
& \nega{$-$.54} & .02 & \nega{$-$.36} & \nega{$-$.34}
& \nega{$-$.50} & .02 & \nega{$-$.33} & \nega{$-$.38}
& \nega{$-$.34} & \nega{$-$.14} & \nega{$-$.10} & \nega{$-$.12} \\
\bottomrule
\end{tabular}
}
\end{table*}

%This systematic underestimation of trace-level PPM approaches may be more consequential than overestimation: planning with too few traces predicted to be active may lead to insufficient capacity to serve more traces, whereas overestimation merely leaves capacity spare, which can be tolerable in domains such as hospital operations. Further, a systematic bias in either direction is worse than no bias at comparable magnitude, since it constitutes a persistent, uncorrected failure mode rather than symmetric error that averages out over time. By this measure, the PPM baselines consistently perform worse than the other approaches, indicating that aggregation of trace-level predictions leads to error propagation: if individual traces are predicted to be shorter than they actually are, the trace-level error may be small (which is optimized during training); however, if many traces are predicted to be too short, the process-level error will be systematic and strong, as our experiments show.

\begin{wrapfigure}{R}{.5\textwidth}
\centering
\vspace{-1em}
\includegraphics[width=\linewidth]{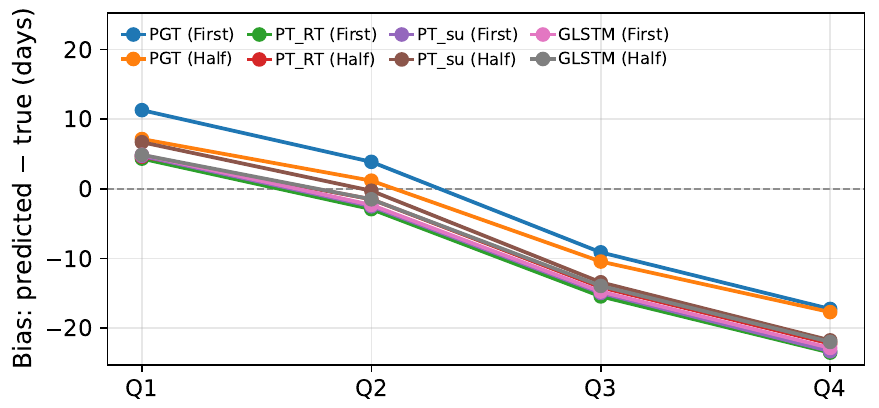}
\vspace{-2.5em}
        \caption{Trace-level bias, grouped into quartiles of actual remaining time (Q1 shortest, Q4 longest).}
        \vspace{-1.3em}
\label{fig:horizon-17}
\end{wrapfigure}

\myparit{Trace-level errors} To identify the sources of process-level errors, we examine the underlying performance on individual traces.\footnote{Note that \textit{Plain} does not allow for trace-level comparisons as newly generated traces cannot be matched to test window traces.} Figure~\ref{fig:horizon-17} shows the prediction error for traces in bpi17-o grouped into quartiles by their horizon: Q1 contains the 25\% of traces with the shortest, Q4 the 25\% with the longest true remaining time. All approaches overestimate the remaining time of traces close to finishing (Q1) and underestimate the remaining time of traces that will run longer (Q3 and Q4), indicating a trend towards predicting the median (boundary of Q2 and Q3). Since underestimation of Q4 is more severe, the overall median NMBE is negative (see \autoref{tab:results_nmbe}). 
This has consequences for CTP and TTP. For TTP, naturally, average throughput times are underestimated if long-running traces finish. For CTP, this error compounds: if a trace is predicted to finish 15 days earlier, CTP will be lower for 15 consecutive days. Thus, the effect is stronger for CTP than for TTP and explains that bad process-level performance of trace-level PPM is due to underestimations of long traces that compound.
We observe the same pattern in all logs, shown in our repository. 

For \textit{Plain} and \textit{First}, this is magnified by a second factor. A trace-level model that sees only one event has a hard time to distinguish longer and shorter traces as the observed prefix gives little information. Thus, it has a disadvantage due to the prediction horizon: it is optimized to predict the short-term future of one trace whereas CTP and TTP aim for long-term developments, which are hard to predict on trace-level without sufficient information on prefixes evolutions.

\mypar{Additional aspects} We also investigated several other aspects of process-level prediction. First, we compared the \textit{training times} of approaches, which show that simulation and trace-level PPM require about 20 times more training time than statistical and ML approaches, which usually are below one minute. Second, we investigated the effect of including \textit{inter-case features} in trace-level PPM to account for interactions between traces. We found that suffix prediction benefits more from those features than remaining time prediction, but neither becomes competitive. Third, we looked at the \textit{sensitivity} of approaches to the amount of available training data by progressively moving the train split to only 20\% of days. We found that this reduction affects performance, but no clear and consistent pattern emerged across approaches and tasks. 
Last, we also investigated the \textit{robustness} of approaches to weight initializations by re-training the models with two more random seeds. Trace-level approaches show variability whereas most process-level approaches are more robust, some even deterministic.
Details on all additional aspects can be found in our repository.

\section{Conclusion}
\label{sec:conclusion}

In this paper, we presented a benchmarking study on process-level prediction. We formalized two process-level prediction tasks and compared statistical, ML, foundation model, and simulation methods with aggregated trace-level predictions. Direct process-level forecasting generally outperforms trace-level PPM. Our findings indicate that trace-level errors, caused by underestimating long-running traces, compound on process-level and lead to systematic errors. 
Even with informational advantage through the provision ground-truth prefixes, trace-level PPM remains largely uncompetitive.
% Providing additional prefix information or inter-case features alleviates this problem only in selected settings and does not make trace-level approaches consistently competitive. Overall, our results suggest that models optimized for individual traces are not necessarily suitable for forecasting aggregate process behavior and that process-level prediction should be treated as a distinct PPM task.

\mypar{Discussion}
Our results provide several broader insights into process-level PPM. First, directly forecasting process-level behavior is more promising than deriving it bottom-up from trace-level predictions, but process-level forecasting itself remains challenging. No approach consistently performs best, and errors increase under complex or changing process behavior, such as trends or combined fluctuations. This highlights the importance of selecting models according to the characteristics of the underlying process. %rather than assuming a universally best forecasting technique.
Second, our results illustrate why good trace-level predictions do not necessarily translate into good process-level predictions. 
%In principle, aggregation could reduce prediction errors if trace-level over- and underestimations were independent and approximately symmetric. 
The applied PPM approaches exhibit systematic rather than symmetric errors: they tend to overestimate short remaining times and, more importantly, underestimate long ones. Such errors compound over time, particularly for CTP, where predicting a trace to finish too early affects every subsequent time window until its actual completion. The observed underestimation of trace-level PPM approaches may be more consequential than overestimation: planning with too few active traces may lead to insufficient capacity, whereas overestimation merely leaves capacity spare, which can be tolerable in domains such as hospitals.
Third, we observe no consistent advantage of remaining-time over suffix prediction, or vice versa, suggesting that the central issue is not the specific prediction target but the mismatch between predicting on trace-level and evaluating aggregate behavior.
This mismatch motivates considering process-level prediction as a distinct PPM task. Models for this task should be trained and evaluated with respect to the process-level KPIs they aim to predict, rather than treating these KPIs as aggregations of trace-level predictions. However, this does not make trace-level PPM obsolete. Trace-level and process-level predictions support complementary types of decisions. The former guides interventions for individual instances, whereas the latter supports decisions concerning capacity, staffing, or process design. Matching the granularity of predictions to the decision they support may thus be an important step toward increasing the practical utility of PPM.

%This systematic underestimation of trace-level PPM approaches may be more consequential than overestimation: planning with too few traces predicted to be active may lead to insufficient capacity to serve more traces, whereas overestimation merely leaves capacity spare, which can be tolerable in domains such as hospital operations. Further, a systematic bias in either direction is worse than no bias at comparable magnitude, since it constitutes a persistent, uncorrected failure mode rather than symmetric error that averages out over time. By this measure, the PPM baselines consistently perform worse than the other approaches, indicating that aggregation of trace-level predictions leads to error propagation: if individual traces are predicted to be shorter than they actually are, the trace-level error may be small (which is optimized during training); however, if many traces are predicted to be too short, the process-level error will be systematic and strong, as our experiments show.

\mypar{Limitations \& Future Work} 
Our study is subject to some limitations and opens several directions for future work. First, we focus only on two process-level KPIs. While they capture important process dimensions, future studies should investigate further KPIs, such as resource utilization, service-level compliance, or cost. Second, we consider daily aggregation and fixed forecasting horizons. Different granularities and horizons may change the relative advantages of approaches and should thus be studied systematically.
Moreover, despite accounting for fade-out periods in real-life logs, our truncation may not always create perfectly steady time series. Also, we only used one arrival rate prediction method for the \textit{Plain} variant of trace-level PPM. While a better arrival rate approach may improve accuracy, the \textit{First} variant results showed that even under perfect information, results hardly improve.
More fundamentally, our results motivate the development of prediction methods that are explicitly optimized on process-level. Such methods may combine trace-level and inter-case information with historical KPI dynamics in hierarchical or multi-task models. Besides this, process-level PPM also requires research on the management side, studying use cases and practitioners' requirements.
Finally, we hope that our study offers a new perspective on PPM and is a step towards increasing its practical adoption.


\mypar{Acknowledgments} Thanks to Daniel Fuhge for his initial work. 
% We acknowledge using DeepL and Claude for code editing. After that, we reviewed and edited the content as needed and take full responsibility for all content. 

\bibliographystyle{ACM-Reference-Format}
\bibliography{sample-base}


%%%%%% Uncomment to show appendix
% \appendix
% \input{appendix}

\end{document}